\documentclass[11pt,a4paper]{article}
\usepackage[hyperref]{emnlp2020}
\usepackage{times}
\usepackage{latexsym}

\usepackage{hyperref}

\usepackage{microtype}
\usepackage{dirtytalk}
\usepackage[ruled, linesnumbered]{algorithm2e}

\aclfinalcopy 

\title{NEU at WNUT-2020 Task 2: Data Augmentation To Tell BERT That \\
Death Is Not Necessarily Informative}

\author{Kumud Chauhan \\
   Northeastern University \\
   \texttt{chauhan.ku@northeastern.edu}
}

\date{}

\begin{document}
\maketitle
\begin{abstract}
Millions of people around the world are sharing COVID-19 related information on social media platforms. Since not all the information shared on the social media is useful, a machine learning system to identify informative posts can help users in finding relevant information. In this paper, we present a BERT classifier system for W-NUT2020 Shared Task 2: Identification of Informative COVID-19 English Tweets. Further, we show that BERT exploits some easy signals to identify informative tweets, and adding simple patterns to uninformative tweets drastically degrades BERT performance. In particular, simply adding \say{10 deaths} to tweets in dev set, reduces BERT F1- score from $92.63$ to $7.28$. We also propose a simple data augmentation technique that helps in improving the robustness and generalization ability of the BERT classifier.    
\end{abstract}

\section{Introduction}
COVID-19 pandemic as well as COVID-19 related information both are spreading across the world rapidly.  Easy access to the internet made the consumption and sharing of information much faster. Millions of people are sharing COVID-19 related information using social media channels such as Facebook, Twitter. Social media is divided over several issues related to masks, social distancing, COVID-19 testing, etc. Unfortunately, the response to the coronavirus has been often determined by people's ideology instead of health officials' guidelines. One challenge with massive information available on social media is to separate useful COVID-19 related information from the noise.  As a testimony, the director of WHO in Munich Security conference said \say{We’re not just fighting an epidemic; we’re fighting an infodemic}\footnote{\url{https://www.who.int/dg/speeches/detail/munich-security-conference}}.

NLP community have taken multiple initiatives to fight this infodemic including misinformation identification ~\cite{shahi2020exploratory}, finding answers to COVID related questions ~\cite{esteva2020co}. 

In this paper, we present our system in the W-NUT 2020 Shared Task 2: Identification of Informative COVID-19 English Tweets \cite{covid19tweet}. We use transformer ~\cite{vaswani2017attention} based models such as BERT ~\cite{devlin2018bert} to classify COVID-19 tweets into informative and uninformative category. We observe that BERT heavily relies on death-related information to identify informative tweets, and can be easily fooled by adding a simple death count information to any uninformative tweet. Further, to improve robustness of the model, we propose a targeted data augmentation technique. Our contributions are as follows:
\begin{itemize}
    \item We show that while BERT based models perform well on the shared task dataset, their performance falls drastically if we add death-related information to uninformative tweets. 
    \item We propose a simple data augmentation method that aims at replacing the most discriminative words with other words to reduce the model's heavy reliance on specific words.
\end{itemize}

\section{Task Description}
The goal of the shared task 2 is to classify English tweets into INFORMATIVE or UNINFORMATIVE category. INFORMATIVE tweets provide information about recovered, suspected, confirmed, and death cases as well as location or travel history of the cases. The dataset consists of 10,000 COVID English Tweets, including 4719 Tweets labeled as INFORMATIVE and 5281 Tweets labeled as UNINFORMATIVE. Table~\ref{dataset} shows the descriptive statistics of the train, dev, and test partition of the dataset.

\begin{table}
\centering
\begin{tabular}{lrl}
\hline \textbf{Split} & \textbf{Informative} & \textbf{Uninformative} \\ \hline
Train & 3303 & 3697 \\
Dev   & 472  & 528 \\
Test  & 944  & 1056\\
\hline
\end{tabular}
\caption{\label{dataset} Train, Dev and Testset statistics}
\end{table}

\section{Method}
The task is formulated as a binary classification task. Since transformers based models provide state of the art performance on the text classification task, we use a pre-trained BERT as a classifier. 

\subsection{Classifiers}
For the classification task, we consider two different variants of the BERT model. Our first model is pre-trained \emph{bert-large-uncased} BERT model ~\cite{devlin2018bert}. The second model is COVID-Twitter-BERT (CT-BERT) which is a transformer-based model pre-trained on a large corpus of Twitter messages on the topic of COVID-19. ~\cite{mller2020covidtwitterbert}. Note that both models are similar in size and use the same base architecture. The only difference is in terms of the data used for pre-training. While \emph{bert-large-uncased} is trained on Wikipedia and Books corpus, CT-BERT is trained on 22.5M tweets corpus. 

\subsection{BERT Exploits Easy Clues For Classification}
 Machine Learning systems can perform well by relying on heuristics that are effective for frequent example types but break down in more challenging cases.~\cite{mccoy2019right, niven2019probing}. In our exploratory experiments, we found that the BERT classifier heavily relies on some easy clues such as the presence of reported deaths related terms such as \say{deaths}, \say{died} to identify INFORMATIVE  tweets. For example, BERT model classify the following uninformative tweet from validation set (1248314498747895813) as informative. 
 
 \textit{Data from National Records of Scotland shows that in the last full week, registered deaths from all causes across Scotland was 60\% higher than the five-year average at 1,741 deaths compared with the average of 1,098." Most of that will be \#COVID19? HTTPURL}

 The above tweet contains death count related information, model fails to recognize it as uninformative tweet. This leads us to experiment with adding such signals to UNINFORMATIVE  tweets, to our surprise, it results in a huge drop in classification performance. For example, simply adding \say{10 deaths} to all tweets in the dev set, reduces BERT F1-score from $92.63$ to $7.28$. 

 Since such models can break for any uninformative tweet which contains variants of reported deaths, it raises questions about models' robustness and usefulness in a real-world application. In this paper, we explore data augmentation as a method to improve the robustness of such models.

\subsection{Data Augmentation To Improve Robustness And Generalization}

\begin{algorithm}[th]
\SetAlgoLined
\SetKwInOut{Input}{Input}
 \Input{ 1. Training Dataset $ \scriptstyle D_{train}$ \\ 2. BERT model}

Find top $N$ most important unigrams $Replacement\_set$ using chi-square test. \\ 
$ \scriptstyle D_{aug} \gets \{\}$ \\
\ForEach{ $ \scriptstyle {\{x_i, y_i\}} \in D_{train}$}{
Replace all word $w \in Replacement\_set$ in $x_i$ with $MASK$ token \\
Generate a new example $\hat{x_i}$ by replacing word $\hat{w}$ using BERT Model such that $\hat{w} \notin Replacement\_set$ \\
$ \scriptstyle D_{aug} \gets D_{aug} \cup {\{\hat{x_i}, y_i\}} $
}
\caption{\label{da_algo} Data Augmentation} 
\end{algorithm}

Data Augmentation have been extensively used to improve classification performance in low-data regime ~\cite{wu2018conditional, kumar2019closer,wei2019eda,kumar2020data}. In our early experiments, we found that pre-trained model-based data augmentation does not improve BERT classification performance in full data regime which is consistent with previous work ~\cite{kumar2019closer}. Instead, we use targeted data augmentation, as described in algorithm \ref{da_algo} to improve the robustness of the BERT classifier. 

Our data augmentation algorithm is motivated by the fact that BERT relies on the presence of simple markers to distinguish between INFORMATIVE vs UNINFORMATIVE classes. We propose to replace such markers with other words so that model does not rely too much on such markers. We hypothesize that such markers should have high discriminative power since the model relies on these words to distinguish between two categories. In our experiments, we use chi-squared stats to find unigrams which have the most discriminative power and use them as markers. Then, we replace such words with $MASK$ token and use BERT to find the replacement of such words. We generate one example for every example in the training set which doubles the size of our training set. We use CT-BERT\textsubscript{Aug} to refer to the CT-BERT model trained using augmented data.

\section{Experiments}
For our experiments, we use Huggingface's transformers package ~\cite{wolf2019huggingface}. We use AdamW optimizer with a learning rate of $4e-5$, batch size of $32$, and $70$ as the max sequence length. We use the pooled representation of the hidden state of the first special token([CLS]) as the sentence embedding. A dropout probability of $0.1$ is applied to the sentence embedding before passing it to the  1-layer Softmax classifier.  All models are trained for $7$ epochs and the best model is selected on the basis of dev set performance.  

For the model robustness experiment, we create a modified dev set where we add \say{10 deaths} to the beginning of all tweets in the dev set and test model performance against that. Since the main purpose of this paper is not to identify such universal adversarial triggers ~\cite{wallace2019universal}, but to highlight the lack of robustness, our modified dev set provides a proxy for robustness experiment.   

\section{Results}
\begin{table}
\centering
\begin{tabular}{lrl}
\hline \textbf{Model} & \textbf{Mean Dev F1} \\ \hline
Bert & 90.46 (0.29)  \\
CT-BERT   & 92.72 (0.18) \\
\hline
CT-BERT\textsubscript{Aug}  & \textbf{92.84} (0.26)  \\
\hline
\end{tabular}
\caption{\label{results} Mean F1 with (STD) on Dev partition of the data. Mean and STD are computed over 5 runs.}
\end{table}

\begin{table}
\centering
\begin{tabular}{ll}
\hline 
\textbf{Top 20 unigrams} \\
\hline 
breaking, bringing, case, cases, confirmed, \\ confirms, county, deaths, department, died, \\ 
employee, help, new, old, positive, \\ 
recovered, reported, tested, total, user \\
\hline
\end{tabular}
\caption{\label{table:unigrams} Top 20 unigram features identified using chi-squared test statistics }
\end{table}

\begin{table}
\centering
\begin{tabular}{lrl}
\hline \textbf{Model} & \textbf{Dev F1} \\ \hline
CT-BERT & 7.28  \\
CT-BERT\textsubscript{Aug}   & 34.68 \\
\hline
\end{tabular}
\caption{\label{table:robustness_results} F1 score on Dev set where \say{10 deaths} is prepended to all tweets in devset}
\end{table}

We use train set to train BERT and CT-BERT model and report performance on dev set. Table~\ref{results} shows the mean F1 performance with STD on dev set. All experiments are repeated 5 times. Since, our task is a tweets classification task, as expected CT-BERT which is pre-trained on tweets data, performs better than the \emph{bert-large-uncased} BERT model. Given that both models have the same number of parameters and have been trained with the same hyperparameters, the better performance of CT-BERT can be attributed to CT-BERT model pre-training. We submitted CT-BERT predictions on the test set for the final submission, and obtained $89.92$ F1 score on the testset\footnote{For final submission, we submitted a CT-BERT model predictions based on dev set accuracy.}. 

Data Augmentation further improves CT-BERT's classification performance from $92.72$ to $92.84$. Table ~\ref{table:unigrams} represents the top 20 unigrams identified using chi-squared statistics \footnote{\url{https://scikit-learn.org/stable/modules/generated/sklearn.feature_selection.chi2.html}} ~\cite{liu1995chi2}. Most of these words appear more frequently in one category over another and that's why these are ranked higher in chi-squared test.      

As shown in Table ~\ref{table:robustness_results}, we observe a sharp drop in CT-BERT performance on dev set when we add \say{10 deaths} to all tweets. This sharp drop comes from that fact that model classified most of the uninformative tweets as informative tweets because of the presence of death count information. While data augmentation helps in improving the robustness, CT-BERT\textsubscript{Aug} still struggles in adversarial setting.     

\section{Discussion}
Pre-trained language models provide state of the art performance on most of the NLP benchmark datasets. Unsurprisingly, CT-BERT does well on the shared task dataset. While these models can be further improved using ensemble techniques, improved fine-tuning ~\cite{sun2019finetune}, a more important research direction is to assess the applicability of such models to identify informative information in real-world applications.

We show the fragile nature of such classifiers using a simple adversarial task where the model fails to classify an uninformative tweet if we add \say{10 deaths} to it. While not reported in the paper, we observe similar results for any \say{N deaths} pattern. On a social media platform such as Twitter, fake news or uninformative tweets might contain a similar pattern, and state of the art classifiers might not be able to correctly identify them. It raises serious concerns about the robustness of such systems. While our proposed data augmentation helps in improving the model robustness, it still falls short from what is expected from a trustworthy system. 

\section{Conclusion}
In this paper, we introduce BERT based classifiers to identify informative tweets. We show that while such classifiers performs well on standard test benchmarks, they exploits easy clues for classification and their performance degrades drastically in the presence of simple adversarial triggers. We show that targeted data augmentation can help in improving the robustness and classification performance of such classifiers. In future, we will explore universal adversarial triggers ~\cite{ song2020universal} to create a more challenging adversarial dataset and will also explore other techniques such as stability training ~\cite{zheng2016improving} to improve model robustness.   

\bibliographystyle{acl_natbib}
\bibliography{emnlp2020}

\begin{thebibliography}{17}
\expandafter\ifx\csname natexlab\endcsname\relax\def\natexlab#1{#1}\fi

\bibitem[{Devlin et~al.(2018)Devlin, Chang, Lee, and
  Toutanova}]{devlin2018bert}
Jacob Devlin, Ming-Wei Chang, Kenton Lee, and Kristina Toutanova. 2018.
\newblock Bert: Pre-training of deep bidirectional transformers for language
  understanding.
\newblock \emph{arXiv preprint arXiv:1810.04805}.

\bibitem[{Esteva et~al.(2020)Esteva, Kale, Paulus, Hashimoto, Yin, Radev, and
  Socher}]{esteva2020co}
Andre Esteva, Anuprit Kale, Romain Paulus, Kazuma Hashimoto, Wenpeng Yin,
  Dragomir Radev, and Richard Socher. 2020.
\newblock Co-search: Covid-19 information retrieval with semantic search,
  question answering, and abstractive summarization.
\newblock \emph{arXiv preprint arXiv:2006.09595}.

\bibitem[{Kumar et~al.(2020)Kumar, Choudhary, and Cho}]{kumar2020data}
Varun Kumar, Ashutosh Choudhary, and Eunah Cho. 2020.
\newblock Data augmentation using pre-trained transformer models.
\newblock \emph{arXiv preprint arXiv:2003.02245}.

\bibitem[{Kumar et~al.(2019)Kumar, Glaude, de~Lichy, and
  Campbell}]{kumar2019closer}
Varun Kumar, Hadrien Glaude, Cyprien de~Lichy, and William Campbell. 2019.
\newblock A closer look at feature space data augmentation for few-shot intent
  classification.
\newblock \emph{arXiv preprint arXiv:1910.04176}.

\bibitem[{Liu and Setiono(1995)}]{liu1995chi2}
Huan Liu and Rudy Setiono. 1995.
\newblock Chi2: Feature selection and discretization of numeric attributes.
\newblock In \emph{Proceedings of 7th IEEE International Conference on Tools
  with Artificial Intelligence}, pages 388--391. IEEE.

\bibitem[{McCoy et~al.(2019)McCoy, Pavlick, and Linzen}]{mccoy2019right}
R~Thomas McCoy, Ellie Pavlick, and Tal Linzen. 2019.
\newblock Right for the wrong reasons: Diagnosing syntactic heuristics in
  natural language inference.
\newblock \emph{arXiv preprint arXiv:1902.01007}.

\bibitem[{M{\"u}ller et~al.(2020)M{\"u}ller, Salath{\'e}, and
  Kummervold}]{mller2020covidtwitterbert}
Martin M{\"u}ller, Marcel Salath{\'e}, and Per~E Kummervold. 2020.
\newblock Covid-twitter-bert: A natural language processing model to analyse
  covid-19 content on twitter.
\newblock \emph{arXiv preprint arXiv:2005.07503}.

\bibitem[{Nguyen et~al.(2020)Nguyen, Vu, Rahimi, Dao, Nguyen, and
  Doan}]{covid19tweet}
Dat~Quoc Nguyen, Thanh Vu, Afshin Rahimi, Mai~Hoang Dao, Linh~The Nguyen, and
  Long Doan. 2020.
\newblock {WNUT-2020 Task 2: Identification of Informative COVID-19 English
  Tweets}.
\newblock In \emph{Proceedings of the 6th Workshop on Noisy User-generated
  Text}.

\bibitem[{Niven and Kao(2019)}]{niven2019probing}
Timothy Niven and Hung-Yu Kao. 2019.
\newblock Probing neural network comprehension of natural language arguments.
\newblock \emph{arXiv preprint arXiv:1907.07355}.

\bibitem[{Shahi et~al.(2020)Shahi, Dirkson, and
  Majchrzak}]{shahi2020exploratory}
Gautam~Kishore Shahi, Anne Dirkson, and Tim~A Majchrzak. 2020.
\newblock An exploratory study of covid-19 misinformation on twitter.
\newblock \emph{arXiv preprint arXiv:2005.05710}.

\bibitem[{Song et~al.(2020)Song, Yu, Peng, and Narasimhan}]{song2020universal}
Liwei Song, Xinwei Yu, Hsuan-Tung Peng, and Karthik Narasimhan. 2020.
\newblock Universal adversarial attacks with natural triggers for text
  classification.
\newblock \emph{arXiv preprint arXiv:2005.00174}.

\bibitem[{Vaswani et~al.(2017)Vaswani, Shazeer, Parmar, Uszkoreit, Jones,
  Gomez, Kaiser, and Polosukhin}]{vaswani2017attention}
Ashish Vaswani, Noam Shazeer, Niki Parmar, Jakob Uszkoreit, Llion Jones,
  Aidan~N Gomez, {\L}ukasz Kaiser, and Illia Polosukhin. 2017.
\newblock Attention is all you need.
\newblock In \emph{Advances in neural information processing systems}, pages
  5998--6008.

\bibitem[{Wallace et~al.(2019)Wallace, Feng, Kandpal, Gardner, and
  Singh}]{wallace2019universal}
Eric Wallace, Shi Feng, Nikhil Kandpal, Matt Gardner, and Sameer Singh. 2019.
\newblock Universal adversarial triggers for attacking and analyzing nlp.
\newblock \emph{arXiv preprint arXiv:1908.07125}.

\bibitem[{Wei and Zou(2019)}]{wei2019eda}
Jason Wei and Kai Zou. 2019.
\newblock Eda: Easy data augmentation techniques for boosting performance on
  text classification tasks.
\newblock \emph{arXiv preprint arXiv:1901.11196}.

\bibitem[{Wolf et~al.(2019)Wolf, Debut, Sanh, Chaumond, Delangue, Moi, Cistac,
  Rault, Louf, Funtowicz et~al.}]{wolf2019huggingface}
Thomas Wolf, Lysandre Debut, Victor Sanh, Julien Chaumond, Clement Delangue,
  Anthony Moi, Pierric Cistac, Tim Rault, R{\'e}mi Louf, Morgan Funtowicz,
  et~al. 2019.
\newblock Huggingface's transformers: State-of-the-art natural language
  processing.
\newblock \emph{ArXiv}, pages arXiv--1910.

\bibitem[{Wu et~al.(2019)Wu, Lv, Zang, Han, and Hu}]{wu2018conditional}
Xing Wu, Shangwen Lv, Liangjun Zang, Jizhong Han, and Songlin Hu. 2019.
\newblock Conditional bert contextual augmentation.
\newblock In \emph{International Conference on Computational Science}, pages
  84--95. Springer.

\bibitem[{Zheng et~al.(2016)Zheng, Song, Leung, and
  Goodfellow}]{zheng2016improving}
Stephan Zheng, Yang Song, Thomas Leung, and Ian Goodfellow. 2016.
\newblock Improving the robustness of deep neural networks via stability
  training.
\newblock In \emph{Proceedings of the ieee conference on computer vision and
  pattern recognition}, pages 4480--4488.

\end{thebibliography}

\end{document}